# *Research Note*
# On the Informativeness of the DNA Promoter Sequences Domain Theory

**Julio Ortega**                                              JULIO@VUSE.VANDERBILT.EDU
*Computer Science Dept., Vanderbilt University*
*P.O. Box 1679, Station B*
*Nashville, TN 37235 USA*

## Abstract

The DNA promoter sequences domain theory and database have become popular for testing systems that integrate empirical and analytical learning. This note reports a simple change and reinterpretation of the domain theory in terms of *M-of-N* concepts, involving no learning, that results in an accuracy of 93.4% on the 106 items of the database. Moreover, an exhaustive search of the space of *M-of-N* domain theory interpretations indicates that the expected accuracy of a randomly chosen interpretation is 76.5%, and that a maximum accuracy of 97.2% is achieved in 12 cases. This demonstrates the informativeness of the domain theory, without the complications of understanding the interactions between various learning algorithms and the theory. In addition, our results help characterize the difficulty of learning using the DNA promoters theory.

## 1. Introduction

The DNA promoter sequences domain theory and database, contributed by M. Noordewier and J. Shavlik to the UCI repository (Murphy & Aha, 1992), have become popular for testing systems that integrate empirical and analytical learning (Hirsh & Japkowicz, 1994; Koppel, Feldman, & Segre, 1994b; Mahoney & Mooney, 1994, 1993; Norton, 1994; Opitz & Shavlik, 1994; Ortega, 1994; Ourston, 1991; Towell, Shavlik, & Noordewier, 1990; Shavlik, Towell, & Noordewier, 1992). The original domain theory, as usually interpreted, is overly specific in that it classifies all of the promoter sequences in the database as negative instances. Since the database consists of 53 positive instances and 53 negative instances, the accuracy over this database is 50%. The learning systems cited above take advantage of the initial domain theory in order to achieve higher accuracy rates, especially with fewer training examples, than the rates achieved by purely inductive methods such as C4.5 and backpropagation. Thus, the informativeness of this theory is acknowledged, despite its 50% accuracy rate using a naive interpretation. However, the extent to which the theory is informative is not easily ascertained; this is implicit in the interactions between each of the learning algorithms and the theory. This note reports a simple change and reinterpretation of the domain theory in terms of *M-of-N* concepts, which involve no learning, that results in an accuracy of 93.4% on the 106 data items. Moreover, an exhaustive search of the space of *M-of-N* interpretations reveals some that achieve 97.2% accuracy.

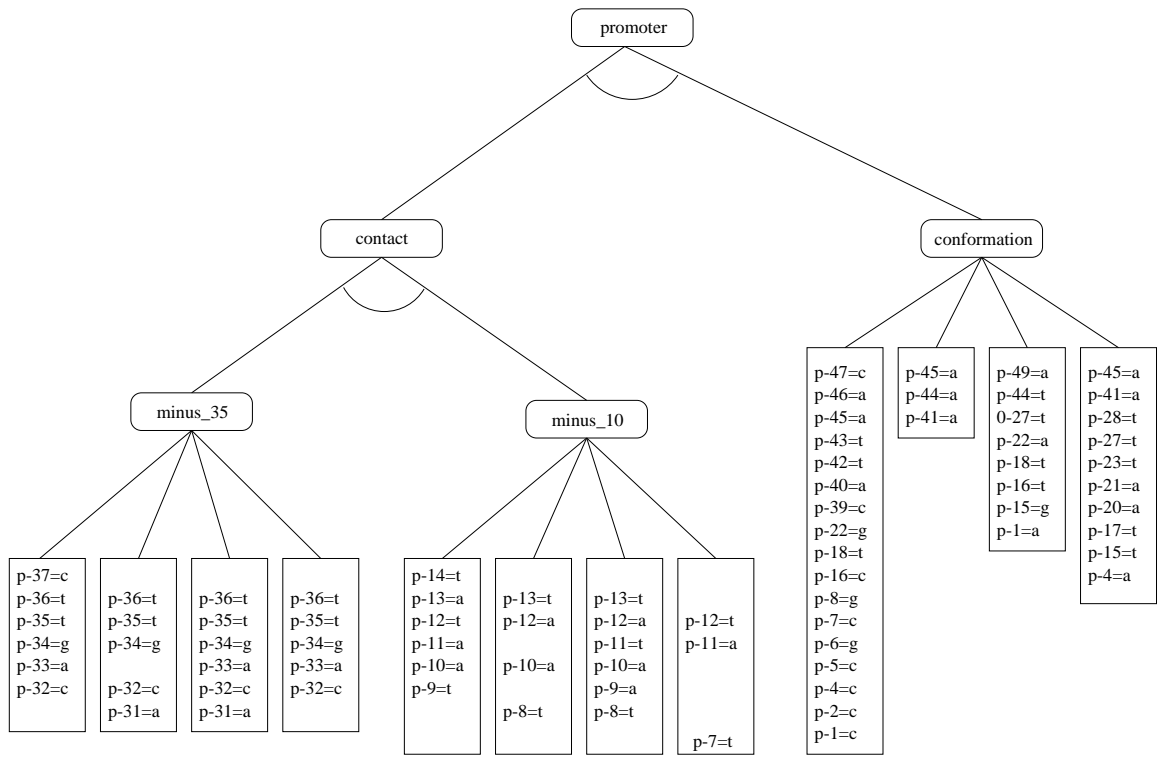

Figure 1: DNA Promoters Theory

## 2. The DNA Promoter Sequences Database and Domain Theory

The sources of the UCI promoter sequences database and domain theory are described by Towell (1990). The rules of the theory were derived from the biological research of O'Neill (1989). The negative examples are contiguous strings from a long DNA sequence believed not to contain promoters. The positive examples of promoters were taken from the compilation of Hawley and McClure (1983). This database has been recently augmented (Harley & Reynolds, 1987; Lisser & Margalit, 1993) and new theories of promoter action are still appearing (Lisser & Margalit, 1993). Nevertheless, the UCI promoter sequences database of examples and domain theory has remained a prominent testbed for evaluating machine learning methods.

The DNA promoters domain theory obtained from the UCI repository is shown in Figure 1 as an AND-OR tree. Each box at a leaf of the tree in Figure 1 is usually interpreted as a conjunction of conditions. Each condition in a box requires that a particular nucleotide appear at a particular position in the sequence. According to this theory, a DNA sequence can be classified as a promoter if two regions of the DNA sequence can be identified. The first region is called a *contact*, and the second a *conformation*. For a conformation region to be identified, any one of the four specific nucleotide sequences shown at the right-hand side of Figure 1 need to be present. For a contact region to be identified, both a *minus_10* and a *minus_35* region also need to be identified. Again, for a *minus_10* or a *minus_35* region to be identified, one of their respective four specific nucleotide sequences need to be present. If a sequence is not classified as a promoter by the domain theory, then that sequence is classified as negative (i.e., not a promoter).

As noted earlier, the promoters domain theory is coupled with a database of 106 items in the UCI repository: 53 examples of promoters and 53 non-examples. When interpreting each leaf of the domain theory as a logical conjunction, the theory classifies all of the data items as negative. Thus, it is clearly too restrictive: No sequence in the database satisfies all the conditions specified in the theory. Two pieces of domain knowledge suggest ways to loosen the conditions of the domain theory. First, the *conformation* condition has very weak biological support. This was implied by the initial KBANN experiments (Shavlik et al., 1992), where none of the learned rules referenced the conformation conditions. In addition, the EITHER system (Ourston, 1991) eliminated rules involving conformation altogether from the domain theory. Eliminating conformation was also supported by a domain expert (Ourston, 1991). The second piece of domain knowledge is that the concepts in this domain tend to take the form of *M-of-N* concepts. Some of the final rules extracted in the KBANN approach take this form. This was also made clear in the NEITHER-MofN system (Baffes & Mooney, 1993), which added a mechanism to handle *M-of-N* concepts to the original learning mechanism of the EITHER/NEITHER system.

## 3. *M-of-N* Interpretations of the DNA Promoters Theory

We can modify the original DNA domain theory as follows, and allow more sequences to be positively classified as promoters: *a)* eliminate the *conformation* condition altogether from the theory, and *b)* reinterpret the conjunctions of conditions in the leaves of Figure 1 as *M-of-N* concepts. As usually interpreted, each of the leaves of Figure 1 is equivalent to a concept of the form *(N-of-N $c_1 c_2 ... c_N$)*. For example, the conjunction *(p-37=c ∧ p-36=t ∧ p-35=t ∧ p-34=g ∧ p-33=a ∧ p-32=c)* of the leftmost leaf is logically equivalent to *(6-of-6 p-37=c p-36=t p-35=t p-34=g p-33=a p-32=c)*.

Progressively less restrictive theories can be created by lowering the number of conditions that need to be satisfied in each leaf of the theory. Thus, a new theory can be constructed where each of the *N-of-N* concepts is substituted by *(N − 1)-of-N*, *(N − 2)-of-N*, etc. The variable $N$ (i.e., the number of conditions at a leaf) is decremented by a constant value $i$ to obtain the value $M$ for each of the *M-of-N* concepts at the leaves of the theory. Figure 2 shows the accuracy of the theories constructed in this manner over all of the examples in the database. As the number of conditions that have to be met for each of the *M-of-N* concepts is lowered, the number of false negatives decreases, and the number of false positive increases. The total number of misclassifications (false negatives plus false positives) is minimized when each leaf is interpreted as a *(N − 2)-of-N* concept, resulting in an accuracy of 93.4%.

Even better accuracies can be obtained if we remove the constant decrement restriction. That is, we allow greater flexibility in choosing different values of $M$ for each of the leaves corresponding to the *minus_35* and *minus_10* regions in Figure 1. By an exhaustive search through all of the 388800 possible combinations of M values we found twelve theories that correctly classify 103 of the 106 examples in the database (i.e., the accuracy of these theories is 97.2%), and 5148 theories of accuracy equal or better than 93.4%. Figure 3 shows the probabilities of obtaining theories of different accuracies when the value of $M$ for each of the *contact* leaves of Figure 1 is chosen at random (but under the restriction that $M \leq N$, where $N$ is the total number of conditions at a particular leaf). The probabilities in Figure 3 were computed by counting the total number of combinations of $M$ values that produced theories of specific accuracies. The mean accuracy of a randomly chosen theory is 76.5%, with a standard deviation of 9.3%.

| M-of-N criteria | Correct Predictions | False Positives | False Negatives | Percent Accuracy |
|---|---|---|---|---|
| N-of-N (original theory) | 53 | 0 | 53 | 50.00% |
| N-of-N w/ conformation rule removed | 57 | 0 | 49 | 53.77% |
| (N − 1)-of-N | 78 | 0 | 28 | 73.58% |
| (N − 2)-of-N | 99 | 1 | 6 | 93.40% |
| (N − 3)-of-N | 90 | 16 | 0 | 84.91% |
| (N − 4)-of-N | 62 | 44 | 0 | 58.49% |

Figure 2: Accuracy of DNA promoters theory under different M-of-N interpretations

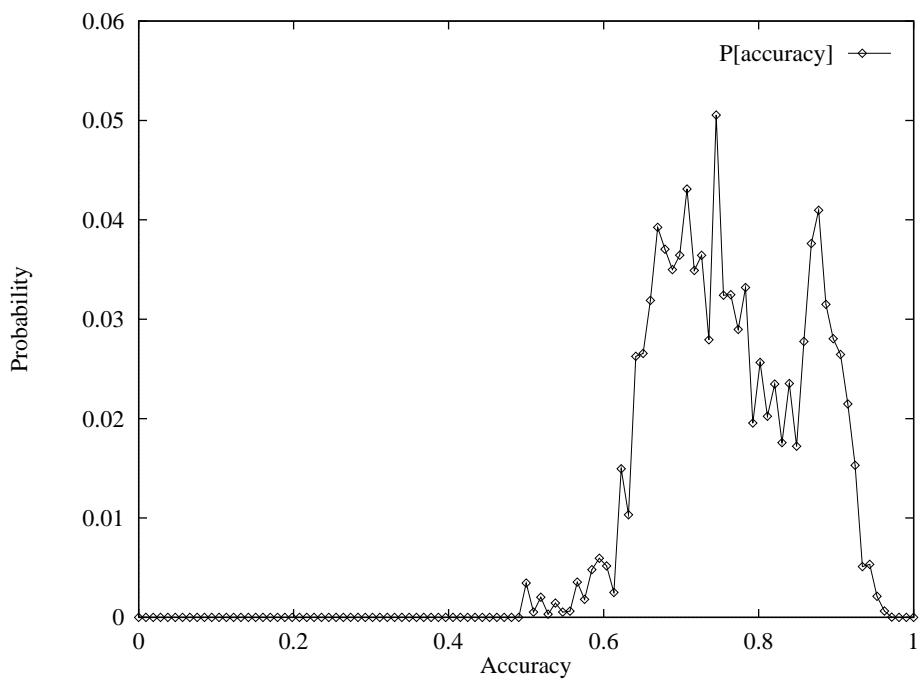

Figure 3: Probability distribution of DNA-theory-interpretation accuracies

These results show that when leaves are interpreted as appropriate M-of-N concepts, the existing DNA domain theory possesses a large amount of predictive information, a fact that has also been pointed out by Koppel et al. (1994a). It is much better than the null power suggested by its initial 50% accuracy, which would be equivalent to random guessing with no theory at all. At the least, the theory allows us to make a single random guess of an M-of-N interpretation with an expected accuracy of 76.5%. As shown in Figure 2 a few random guesses allow us to do much better than that.

## 4. Learning with the DNA Promoters Theory

The accuracies of various systems that integrate analytical and empirical learning are around 93% (Baffes & Mooney, 1993). These results are typically means computed over multiple trials with 80-85 training examples and 21-26 tests examples. Our reported accuracies of 93.4% and 97.2% are not based on such splits of training and test data. Instead, they represent the maximum accuracies (relative to the database of 106 examples) that could be obtained by learning algorithms with certain representational biases. For example, 93.4% is the maximum accuracy that may be achieved by a learning system that identifies *(N − i)-of-N* concepts at the leaves, where $i$ is constant across leaves. This approach can be converted into a learning task where the learner identifies the optimal value of $i$ given a set of training examples, and evaluates the resultant classifier using a test set. The results of this algorithm, averaged over 100 trials, produce mean accuracies of 88.7% with 10 training examples, and 92.5% with 85 training examples (on a test set of 21 examples). These results are similar to the best of the algorithms reported by Baffes and Mooney (1993).

Koppel et al. (1994a) also show that there is considerable information in a "reinterpreted" promoters domain theory. In their DOP (Degree of Provedness) classification methodology, the logical operations of a propositional domain theory (AND, OR, or NOT) are replaced by arithmetic equivalents that contain a degree of uncertainty. Rather than directly returning a truth value indicating whether an example is positive, the system first calculates the DOP numerical score for that example. If the DOP score value is greater than a pre-specified threshold value, then the example is considered "sufficiently" proved and thus classified as a positive example by the theory. Otherwise, the example is classified as negative. Koppel et al. determine the threshold value with two pieces of knowledge: a) the distribution of the DOP score over all examples, and b) the proportion ($n\%$) of positive examples in the database. The DOP values for all examples are sorted, so that the threshold value can be set to the value that separates the $n\%$ of the examples with highest DOP values from the rest. An important assumption is that the domain theory be of a certain proof-additive nature, so that DOP values will be higher for positive examples than for negative examples. The DOP classification methodology achieves a high accuracy (92.5%) when applied to the DNA promoter sequences domain theory. As with our approach, this accuracy is not based on a split of the available data into training and test sets, and represents an upper bound on the accuracy that could be obtained if their method was converted into a learning algorithm. The DOP classification methodology could be converted into a learning algorithm by estimating the distribution of DOP values over a set of training examples.

## 5. Concluding Remarks

This note does not detail a new learning algorithm. Rather, it demonstrates that a suitable learning model in the promoters domain is finding the correct number, $M$, for each of the *M-of-N* concepts at the leaves of the original domain theory.[1] Assessing the difficulty of learning using the available theory is usually complicated by the need to understand the learning algorithms that exploit the theory. The theory-accuracy distribution of Figure 3 helps characterize learning complexity in this domain (under the *M-of-N* model) and provides a dimension along which to evaluate the performance of learning algorithms that use the DNA promoter's theory as their testbed.

---

1. Some algorithms may introduce some structural modification to the theory (i.e., add/delete clauses and conditions). However, the increase in accuracy due to these structural modifications is negligible in the case of the promoters domain, as illustrated by the high accuracies that can be obtained without them.

## Acknowledgements

This research was supported by a grant from NASA Ames Research Center (NAG 2-834) to Doug Fisher. I am grateful for the suggestions of Doug Fisher, Stefanos Manganaris, Doug Talbert, Jing Lin, as well as comments and pointers from Larry Hunter and anonymous JAIR referees.